\documentclass{article}
\usepackage{spconf,amsmath,amssymb,booktabs,multirow,graphicx,url,stfloats}
\usepackage[hidelinks]{hyperref}
\usepackage{microtype}
\usepackage{enumitem}
\usepackage{balance}

\graphicspath{{figures/}}
\setlist[itemize]{leftmargin=*,nosep,topsep=1pt}

\newcommand{\ESR}{\textsc{ESR}}
\newcommand{\EOR}{\textsc{Exp.-0/1-Risk}}
\newcommand{\EIG}{\textsc{E-Entropy}}

\title{Beyond Information Seeking: Severity-Aware\\
Question Supervision for Proactive Medical Dialogue}

\name{
\begin{tabular}{c}
Chenxuan Li$^{1}$ \qquad
Xinrong Chen$^{1}$ \qquad
Luyan Zhang$^{2}$ \qquad
Peidong Jia$^{1}$\\
Runfan Zheng$^{4}$ \qquad
Zhongyu Zhao$^{1}$ \qquad
Xuecheng Shang$^{1}$ \qquad
Peixing Wan$^{3}$\sthanks{Corresponding author: peixing@bjmu.edu.cn}
\end{tabular}
}

\address{
$^{1}$Peking University, Beijing, China \qquad
$^{2}$Northeastern University, Boston, MA, USA\\
$^{3}$Department of Medical Bioinformatics,
School of Basic Medical Sciences,\\
Peking University, Beijing, China\\
$^{4}$Sun Yat-sen University, Guangzhou, China
}

\begin{document}
\ninept
\maketitle


\begin{abstract}
Proactive medical dialogue requires an agent to decide what to ask from incomplete patient information. Existing information-seeking approaches commonly prioritize questions that most reduce diagnostic uncertainty, but this criterion overlooks an important property of medical diagnosis: different diagnostic errors can carry substantially different consequences. The most informative question may therefore differ from the one most valuable for the downstream decision. We propose Expected-Severity-Risk (ESR), a consequence-aware question-supervision objective that values each candidate by its expected reduction in severity-aware terminal risk. Because questions must be selected before their answers are observed, ESR marginalizes over possible answers using train-only population statistics. Its rankings are then distilled into a prefix-only language policy, requiring no teacher-side risk computation at deployment. Across three matched Qwen3-4B training seeds on DDxPlus, ESR reduces mean high-severity diagnostic miss from 0.0645 to 0.0455 (29.5\% relative reduction) and improves mean diagnostic accuracy from 0.9123 to 0.9320 while requiring only 0.14 additional questions per dialogue. Fixed-budget analyses show that the distinction persists when question count is controlled, while a matched expected-0/1-risk student control further isolates the contribution of asymmetric severity weighting. These results support moving proactive medical dialogue beyond uncertainty reduction toward consequence-aware evidence acquisition.
\end{abstract}



\section{Introduction}

Clinical diagnosis is inherently interactive: decision-relevant
evidence is progressively acquired through targeted questions.
Proactive medical dialogue asks language agents to make the same
acquisition decision, choosing what to ask next from a partial patient
history before the answer is known. Early systems used dialogue to
acquire additional symptoms for diagnosis
\cite{wei2018task,xu2019knowledge,luo2020knowledge}.
Representative approaches include information-gain acquisition
\cite{mackay1992information,ma2019eddi,wong2026medclarify},
cost-sensitive active feature acquisition
\cite{greiner2002cost,saar2009active,shim2018joint,li2021active},
and RL-based questioning policies
\cite{feng2026doctoragent,ding2026promed,lai2026doctorr1,cao2026atpo,chu2025qwenlookagainguiding,chen2026residualdecodingmitigatinghallucinations}.

Information-based objectives provide a natural selection-time
criterion, but posterior concentration is only a proxy for downstream
decision quality. In medical diagnosis, errors can have substantially
different consequences: a question separating common alternatives may
reduce uncertainty more than one that helps rule out a less likely but
more severe condition. Classical cost-sensitive acquisition already
uses expected classification cost, often together with acquisition
cost, including in sequential medical diagnosis
\cite{greiner2002cost,ji2007cost}. Our contribution is therefore not
expected-risk acquisition itself, but its adaptation to selection-time
question supervision for language policies when candidate answers are
unobserved. Reinforcement learning can optimize dialogue-level
outcomes, yet question value is then entangled with policy
optimization, reward design, and resulting trajectories. This
motivates a more direct question:
\emph{can evidence be valued by downstream diagnostic consequence
using only the information available at question-selection time?}

\begin{figure*}[t]
    \centering
    \includegraphics[width=0.96\textwidth]{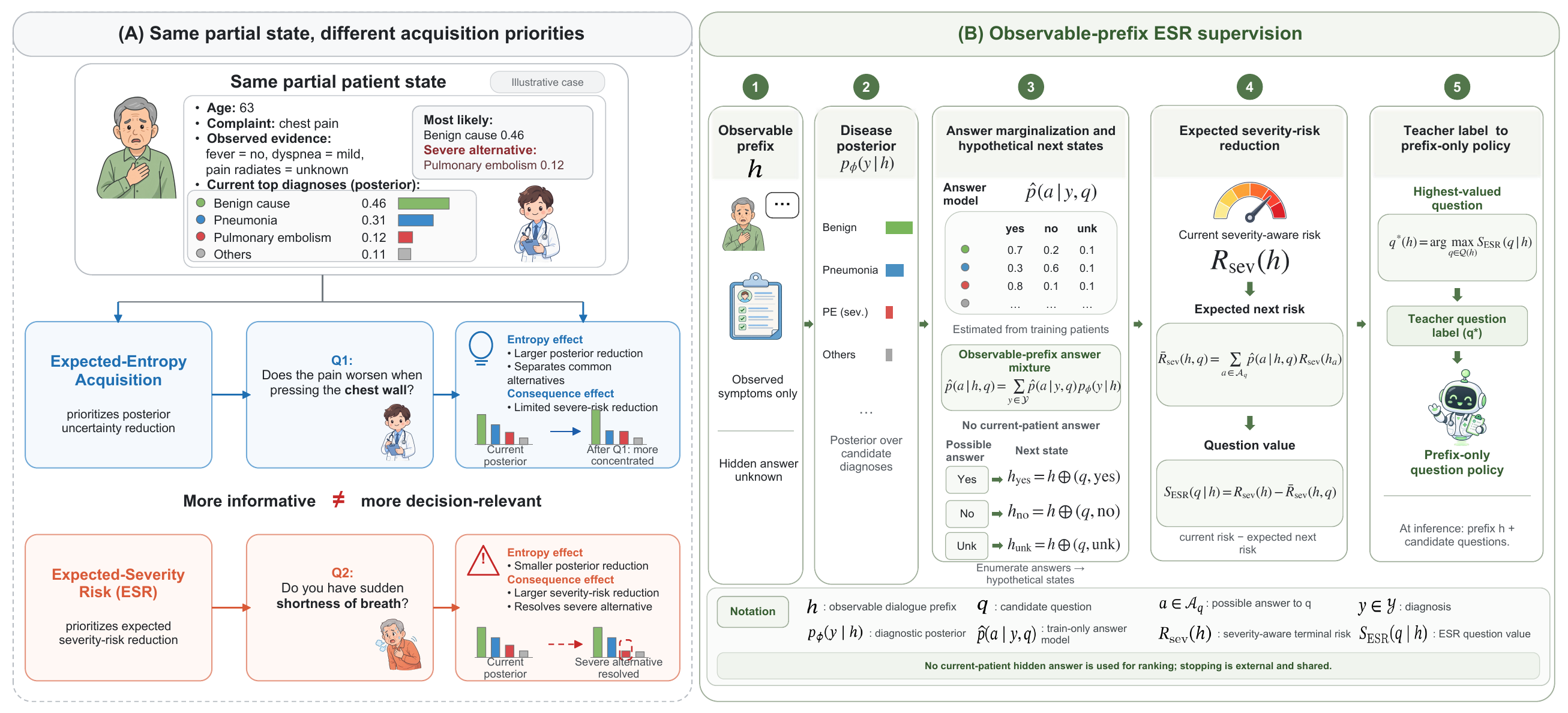}
    \caption{\textbf{Information value and diagnostic consequence can prioritize different questions.}
    \ESR{} evaluates candidate questions before their answers are
    observed and prioritizes evidence according to expected reduction
    in severity-aware terminal risk. The current patient's hidden answer
    is not used during question ranking.}
    \label{fig:concept}
\end{figure*}

We introduce Expected-Severity-Risk (\ESR{}) to operationalize this
idea. As illustrated in Fig.~\ref{fig:concept}, a fixed downstream
diagnostic model defines terminal decision risk, while disease
severity provides an asymmetric consequence signal. Because answers
are unobserved at selection time, \ESR{} marginalizes over possible
answers and scores each question by its expected reduction in
severity-aware terminal risk. The rankings are distilled into a prefix-only policy, with no teacher computation at deployment.

Across three matched Qwen3-4B training seeds on DDxPlus, \ESR{}
reduces mean high-severity diagnostic miss from .0645 to .0455
($-29.5\%$) and improves diagnostic accuracy from .9123 to .9320
with only 0.14 additional questions per dialogue. Fixed-budget
analyses show that the acquisition objectives remain behaviorally
distinct when question count is controlled, while a matched distilled
\EOR{} control isolates the benefit of severity-aware weighting over
expected 0/1-risk supervision. Together, these results support
decision-aware question supervision beyond purely information-seeking
acquisition.

\textbf{Contributions.}
Our contributions are threefold:
\begin{itemize}

\item We propose \ESR{}, a consequence-aware question-supervision
objective that combines severity-aware terminal risk with
selection-time marginalization over unobserved answers.

\item We distill these rankings into a prefix-only language policy and
use matched student controls to isolate severity-aware weighting from
expected 0/1-risk supervision.

\item Across three Qwen3-4B training seeds, \ESR{} reduces mean
high-severity diagnostic miss by 29.5\% and improves diagnostic
accuracy from .9123 to .9320 with only 0.14 additional questions per
dialogue.

\end{itemize}


\section{Decision-Aware Question Supervision}
\subsection{Problem Formulation}

We study proactive diagnosis as sequential evidence acquisition under partial observation. At turn $t$, the dialogue prefix $h_t$ contains the patient evidence observed so far, while $\mathcal{Q}_t$ denotes the unanswered candidate questions. The policy selects $q_t\in\mathcal{Q}_t$ before observing its answer
$a_t$, after which the dialogue state becomes
\begin{equation}
h_{t+1}=h_t\oplus(q_t,a_t).
\end{equation}

A fixed diagnostic model maps each partial state to a disease posterior $p_\phi(y\mid h_t)$ and ultimately supports a terminal diagnosis. We treat this model as the downstream decision maker and focus on
which available question is most valuable to acquire next from the
current partial state.

Thus, question value must account for asymmetric diagnostic
consequences and uncertainty over candidate answers that are
unobserved at selection time.

\subsection{Expected-Severity-Risk Question Supervision}
Motivated by these two requirements, we construct \ESR{} as a selection-time teacher that connects question acquisition to the downstream diagnostic decision. It first assigns a consequence-aware risk to each partial evidence state, then evaluates a candidate question over the possible next states induced by its unobserved answer.

\textbf{Severity-aware terminal risk.}
We first quantify the downstream consequence of a decision made from a partial evidence state. DDxPlus \cite{tchango2022ddxplus} assigns each disease a severity level $r(y)$, with smaller values indicating greater severity. We normalize it as
\begin{equation}
w(y)=
\frac{r_{\max}-r(y)}
     {r_{\max}-r_{\min}},
\label{eq:severity}
\end{equation}
and define uniform and severity-aware terminal losses:
\begin{equation}
\begin{aligned}
C_{01}(d,y)
&=\mathbb{I}[d\neq y],\\
C_{\rm sev}(d,y)
&=\mathbb{I}[d\neq y]\bigl(1+2w(y)\bigr),
\qquad d\in\mathcal{Y},\\
C_{\ell}(\bot,y)
&=c_\bot=0.5,
\qquad \ell\in\{01,{\rm sev}\}.
\end{aligned}
\label{eq:loss}
\end{equation}

Given the diagnostic posterior, the terminal risk of state $h$ is
\begin{equation}
R_{\ell}(h)
=
\min_{d\in\mathcal{Y}\cup\{\bot\}}
\sum_{y\in\mathcal{Y}}
p_\phi(y\mid h)C_{\ell}(d,y).
\label{eq:risk}
\end{equation}
$R_{01}$ treats all diagnostic errors equally, whereas
$R_{\rm sev}$ gives greater consequence to errors involving more severe conditions. Thus, terminal risk characterizes the downstream decision induced by the current evidence, rather than only how concentrated the disease posterior is.

\textbf{Selection-time answer model.}
Terminal risk evaluates an evidence state, whereas asking a question can produce different next states depending on the patient's answer. Because that answer is unavailable when $q$ is selected, we represent these possible outcomes with a predictive answer distribution.

We estimate a disease-conditioned categorical model
$\widehat p(a\mid y,q)$ from the training population using add-one smoothing. Combining it with the current disease posterior gives
\begin{equation}
\widehat p(a\mid h,q)
=
\sum_{y\in\mathcal{Y}}
\widehat p(a\mid y,q)\,p_\phi(y\mid h).
\label{eq:answer}
\end{equation}
This distribution weights the hypothetical next states
$h\oplus(q,a)$ that may result from asking $q$, using information available before the actual answer is observed.

\textbf{Expected question value.}
We can now value a candidate by averaging the terminal risk of its possible next states. Expected-Severity-Risk is defined as
\begin{equation}
S_{\rm ESR}(q\mid h)
=
R_{\rm sev}(h)
-
\mathbb{E}_{a\sim\widehat p(\cdot\mid h,q)}
\left[
R_{\rm sev}\!\left(h\oplus(q,a)\right)
\right].
\label{eq:score}
\end{equation}
A high score indicates that asking $q$ is expected to move the downstream diagnostic decision toward lower severity-aware risk.

For controlled comparison, \EIG{} replaces terminal risk with posterior entropy while retaining the same answer marginalization, whereas \EOR{} replaces $R_{\rm sev}$ with uniform risk $R_{01}$.
Thus, \EIG{} provides the information-seeking reference, while the \EOR{}--\ESR{} contrast isolates asymmetric severity weighting.

\subsection{Distillation into a Prefix-Only Dialogue Policy}

We distill the teacher rankings into a prefix-only language policy:
for each observable prefix and candidate set, the highest-valued
question becomes the next-question label. At inference, the student
selects directly from the observable context without access to
$p_\phi$, the answer model, hidden answers, or teacher utilities.


\section{Experiments}

\subsection{Experimental Setup}

\textbf{Environment and evaluator.}
We construct a nine-condition DDxPlus subset
\cite{tchango2022ddxplus} from training-set differential diagnoses.
A normalized pathology co-occurrence graph is built from 250,000
training cases, and a dense connected cluster of nine overlapping
conditions is selected greedily and frozen before sampling 5,000
training, 800 validation, and 1,000 held-out test patients. Dialogues
start from the chief complaint and acquire evidence from a fixed
50-question inventory. A frozen multinomial logistic classifier over
structured partial-state features provides the disease posterior and
terminal diagnosis; its logits are temperature-scaled by validation
NLL ($T=0.834$) \cite{guo2017calibration}. The same evaluator is
used for every acquisition objective.

\textbf{Matched student training.}
The \EIG{} and \ESR{} students share the same Protocol-SFT Qwen3-4B
initialization \cite{yang2025qwen3} and 17,813 prefix--candidate
training states. We use rank-16 LoRA \cite{hu2022lora,chen-etal-2025-guilomo}, learning rate
$10^{-4}$, effective batch size 16, and three epochs, and repeat the
main comparison with seeds 42, 43, and 44. Within each seed, training
inputs and optimization settings are identical across objectives;
only the teacher-derived next-question target differs.
Candidate-constrained decoding restricts both students to the same
available questions.

\textbf{Interaction protocol.}
Both policies use the same candidate interface and external stopping
rule. A dialogue terminates when the minimum-risk action under
$C_{\rm sev}$ is a diagnosis and $R_{\rm sev}(h)\leq0.20$;
otherwise the student selects among the first 15 unanswered questions
in the fixed 50-question inventory. The primary interaction budget is
$B=15$. If the stopping criterion is not met at the hard budget, the
same evaluator returns its minimum-risk diagnosis or abstention;
cap-censored cases remain included in all terminal metrics.

\textbf{Metrics.}
Primary outcomes are accuracy, average question count (Q), and
high-severity conditional miss (HSM). The severity threshold is fixed
before test evaluation; 667 of the 1,000 test patients satisfy
$w(y)\geq0.5$:
\begin{equation}
\mathrm{HSM}=\Pr(d_T\neq y\mid w(y)\geq0.5).
\end{equation}
Secondary measures include severity-weighted error, population
high-severity error, hard-cap censoring, and trajectory cost
$J=C_{\rm sev}(d_T,y)+0.03N_q$.
Main results are mean$\pm$SD across training seeds; paired 95\% CIs
use 2,000 patient bootstrap resamples within each seed.

\begin{table*}[t]
\caption{\textbf{Main shared-stopping comparison across three seeds.}
Values are mean$\pm$SD over seeds 42/43/44; only next-question
supervision differs. $\Delta$ is the seed-wise
\ESR{}$-$\EIG{} difference.}
\label{tab:main}
\centering
\setlength{\tabcolsep}{1.8pt}
\begin{tabular}{lccccccc}
\toprule
Method &
Acc.$\uparrow$ &
Q &
HSM$\downarrow$ &
Sev.$\downarrow$ &
Pop.-HS$\downarrow$ &
Cost$\downarrow$ &
Cap$\downarrow$\\
\midrule
\EIG{} &
$.9123\!\pm\!.0015$ &
$2.182\!\pm\!.020$ &
$.0645\!\pm\!.0000$ &
$.0447\!\pm\!.0003$ &
$.0430\!\pm\!.0000$ &
$.2377\!\pm\!.0014$ &
$.0073\!\pm\!.0015$\\
\textbf{\ESR{}} &
$\mathbf{.9320\!\pm\!.0017}$ &
$2.322\!\pm\!.025$ &
$\mathbf{.0455\!\pm\!.0048}$ &
$\mathbf{.0333\!\pm\!.0015}$ &
$\mathbf{.0303\!\pm\!.0032}$ &
$\mathbf{.1886\!\pm\!.0008}$ &
$.0177\!\pm\!.0038$\\
\midrule
$\Delta$ (\ESR{}$-$\EIG{}) &
$+.0197\!\pm\!.0025$ &
$+.140\!\pm\!.015$ &
$-.0190\!\pm\!.0048$ &
$-.0113\!\pm\!.0019$ &
$-.0127\!\pm\!.0032$ &
$-.0491\!\pm\!.0021$ &
$+.0103\!\pm\!.0051$\\
\bottomrule
\end{tabular}
\end{table*}

\subsection{Main Results}

The primary comparison asks whether changing only the question-value
supervision changes downstream diagnosis under otherwise matched
training and interaction. Table~\ref{tab:main} shows a consistent
shift toward a better high-severity error profile.

Across three matched seeds, \ESR{} reduces mean HSM from
$.0645\!\pm\!.0000$ to $.0455\!\pm\!.0048$ ($-29.5\%$), while
mean accuracy increases from $.9123\!\pm\!.0015$ to
$.9320\!\pm\!.0017$. This change requires only
$0.140\!\pm\!.015$ additional questions on average.

Patient-level uncertainty supports the same overall pattern. Paired
95\% CIs for $\Delta$Acc exclude zero in all three seeds:
$[.005,.036]$, $[.007,.038]$, and $[.002,.032]$. For
$\Delta$HSM, the intervals exclude zero for seeds 42 and 43
($[-.0398,-.0030]$ and $[-.0411,-.0030]$) and overlap zero for
seed 44 ($[-.0319,.0059]$).

Objective-aligned secondary measures move consistently with the
primary result: severity-weighted error decreases from .0447 to
.0333, population high-severity error from .0430 to .0303, and proxy
cost from .2377 to .1886. Together, the main comparison indicates
that changing question supervision alters the downstream
accuracy--severity trade-off rather than merely reproducing the same
diagnostic behavior.

\subsection{Information and Decision Value Prefer Different Evidence}
Improved terminal outcomes alone do not show that \ESR{} represents a
different notion of evidence value: different early questions can
simply lead the two policies to different later states. We therefore
compare the teacher objectives on 2,000 identical partial patient
states with identical candidate sets, removing trajectory history as
an explanation.

The objectives exhibit a crossed preference. Questions selected by
\EIG{} yield larger mean entropy reduction than those selected by
\ESR{} (.2316 versus .1953), whereas \ESR{} choices yield larger
mean severity-risk reduction (.0902 versus .0810). In 11.4\% of
states, the conflict is explicit: \EIG{} selects the question with
greater uncertainty reduction while \ESR{} selects a different
question with greater severity-risk reduction.

Thus, the distinction appears before different dialogue trajectories
have developed. At the same patient state, the two objectives can
disagree on which evidence is worth acquiring because they optimize
different notions of value. This is the information--decision
mismatch illustrated in Fig.~\ref{fig:concept}.

\begin{figure*}[t]
  \centering
  \includegraphics[width=0.90\textwidth]{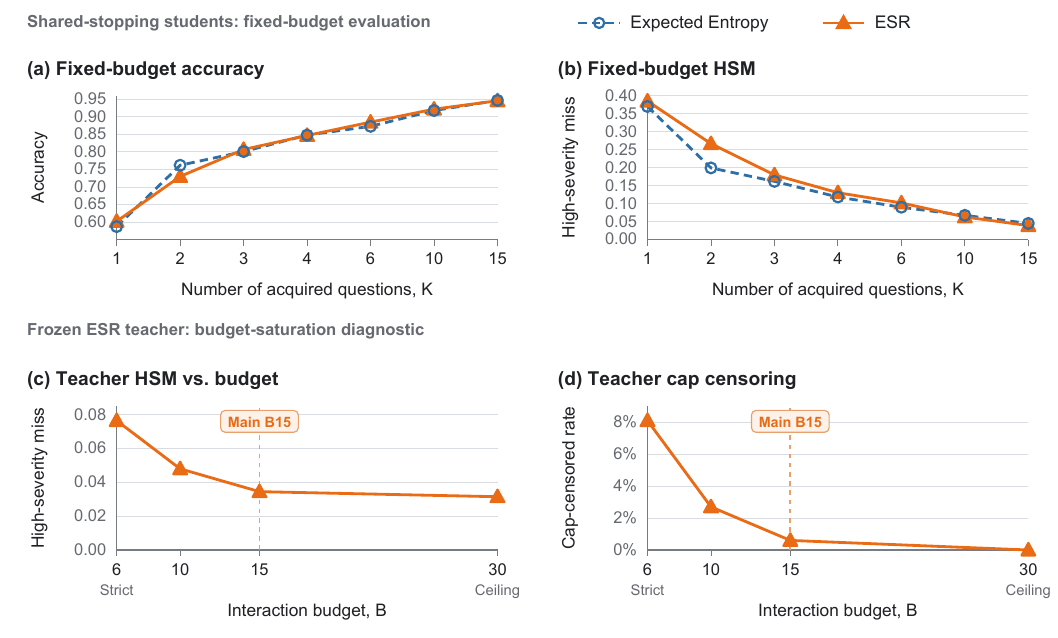}
  \caption{\textbf{Question ranking across interaction budgets.}
  Panels (a--b) compare the frozen seed-42 students under identical
fixed question budgets. Panels (c--d) characterize frozen-teacher
  performance and censoring across horizons.}
  \label{fig:budget}
\end{figure*}

\subsection{Controlling for Dialogue Length}

The shared-stopping comparison leaves a simple alternative
explanation: \ESR{} asks 0.14 more questions on average and might
benefit merely from acquiring additional evidence. To separate
question ranking from dialogue length, we disable stopping and force
the frozen seed-42 \EIG{} and \ESR{} students to acquire exactly
$K$ questions.

The effect is budget dependent rather than uniformly favorable to
\ESR{}. At $K=1$, \ESR{} has slightly higher accuracy (.602 versus
.587) but higher HSM (.385 versus .370). At $K=10$, \ESR{} reaches
.922 accuracy and .0630 HSM, compared with .918 and .0675 for
\EIG{}. At $K=15$, accuracy is effectively matched
(.946 versus .947), while \ESR{} retains lower HSM
(.0390 versus .0450).

Thus, entropy remains competitive at short horizons, whereas
consequence-aware ranking produces a more favorable high-severity
error profile at larger budgets. The shared-stopping horizon $B=15$
was fixed before the current student comparison; the frozen-teacher
sweep in Fig.~\ref{fig:budget}(c--d) reduces censoring from .081 at
$B=6$ to .006 at $B=15$, while extending to $B=30$ changes HSM only
from .0345 to .0315.

\subsection{Attributing the Effect to Severity Weighting}

Having established that the objectives can prefer different evidence,
we next ask whether severity weighting itself is responsible for the
high-severity behavior. A generic decision-aware objective might
already be sufficient even if all diagnostic errors are treated
equally.

We test this alternative using matched frozen seed-42 \EOR{} and
\ESR{} students. Both use the same answer marginalization, training
inputs, initialization, optimization, candidate interface, and
stopping rule; only the teacher valuation changes from expected
0/1 risk to severity-aware risk.

\begin{table}[!h]
\caption{\textbf{Frozen seed-42 student objective decomposition.}
All methods use expected-answer marginalization.}
\label{tab:component}
\centering
\setlength{\tabcolsep}{1.8pt}
\begin{tabular}{lcccc}
\toprule
Method & Acc.$\uparrow$ & Q & HSM$\downarrow$ & Sev.$\downarrow$\\
\midrule
\EIG{} & .911 & 2.161 & .0645 & .0447\\
\EOR{} & .913 & 2.509 & .0690 & .0447\\
\textbf{\ESR{}} &
\textbf{.931} & 2.294 & \textbf{.0435} & \textbf{.0337}\\
\bottomrule
\end{tabular}
\end{table}

The distinction survives policy distillation. Replacing expected
0/1 risk with severity-aware risk increases accuracy from .913 to
.931 and reduces HSM from .0690 to .0435, while question count also
decreases from 2.509 to 2.294. Paired patient-level 95\% CIs for the
\ESR{}$-$\EOR{} contrast are $[.002,.034]$ for accuracy and
$[-.0441,-.0073]$ for HSM.

Together, this matched student contrast provides direct evidence that
asymmetric severity weighting, beyond generic expected classification
risk, redirects the learned acquisition policy toward a more favorable
high-severity error profile.


\section{Conclusion}

We introduced \ESR{}, a selection-time objective that values medical
questions by their expected reduction in severity-aware diagnostic
risk. On a training-derived high-overlap DDxPlus disease cluster,
\ESR{} reduces mean high-severity diagnostic miss by 29.5\% across
three matched Qwen3-4B training seeds while improving accuracy with
only a small increase in question count. Same-state analysis shows
that information and decision value can prioritize different evidence,
while a matched distilled 0/1-risk control identifies asymmetric
severity weighting as a key contributor to the improved high-severity
profile. These results establish consequence-aware question
supervision as a practical approach to aligning proactive evidence
acquisition with downstream diagnostic consequence.


\balance
\section{Compliance with Ethical Standards}
This study uses the publicly available synthetic DDxPlus benchmark
and involves no human participants or identifiable patient data.
No ethical approval was required. The authors declare no conflicts
of interest.

\bibliographystyle{IEEEbib}
\bibliography{refs}

\end{document}